\documentclass[10pt,twocolumn,letterpaper]{article}

\usepackage{cvpr}
\usepackage{times}
\usepackage{epsfig}
\usepackage{graphicx}
\usepackage{amsmath}
\usepackage{amssymb}

\usepackage[breaklinks=true,bookmarks=false]{hyperref}

\cvprfinalcopy 

\def\cvprPaperID{****} 

\begin{document}

\title{Fully Convolutional Network for Automatic Road Extraction from Satellite Imagery}

\author{Alexander V. Buslaev\\
Mapbox R$\&$D Center\\
Minsk, 220030, Belarus\\
{\tt\small aleksandr.buslaev@mapbox.com}
\and
Selim S. Seferbekov\\
Veeva Systems\\
Frankfurt am Main, 60314, Germany\\
{\tt\small selim.seferbekov@veeva.com}
\and
Vladimir I. Iglovikov\\
Level5 Engineering Center, Lyft Inc\\
Palo Alto, CA 94304, USA\\
{\tt\small iglovikov@gmail.com}
\and
Alexey A. Shvets\\
Massachusetts Institute of Technology\\
Cambridge, MA 02139, USA\\
{\tt\small shvets@mit.edu}
}

\maketitle

\begin{abstract}
    Analysis of high-resolution satellite images has been an important research topic for traffic management, city planning, and road monitoring. One of the problems here is automatic and precise road extraction. From an original image, it is difficult and computationally expensive to extract roads due to presences of other road-like features with straight edges. In this paper, we propose an approach for automatic road extraction based on a fully convolutional neural network of U-net family. This network consists of ResNet-34 pre-trained on ImageNet and decoder adapted from vanilla U-Net. Based on validation results, leaderboard and our own experience this network shows superior results for the DEEPGLOBE - CVPR 2018 road extraction sub-challenge. Moreover, this network uses moderate memory that allows using just one GTX 1080 or 1080ti video cards to perform whole training and makes pretty fast predictions.  
\end{abstract}

\section{Introduction}
High-resolution satellite imagery opens new possibilities for the extraction of linear features such as roads \cite{wang2016review}. The advantages of this data compared to aerial imagery are the almost worldwide availability, and sometimes the imagery data contains additional spectral channels. The spatial resolution with 0.5-1.0 meters is worse than for aerial imagery, but for the road extraction, it is sufficient \cite{deepglobe_website, demir2018deepglobe}. The worldwide availability of the data makes it possible to produce topographic databases for nearly any region of the earth. It, in turn, can help in various industries to enhance their productivity and quality of work whether it is for military purposes of disaster prevention or relief.

Reliable image segmentation is one of the important tasks in computer vision. Semantic image segmentation essentially involves dividing images into meaningful regions, which can be viewed as a pixel level classification task. The most straightforward (and slow) approach to such problem is manual segmentation of the images. However, this is a time-consuming process that is prone to mistakes and inconsistencies that are unavoidable when human data curators are involved. To overcome this bottleneck, automatic means are needed. Automating the treatment provides a systematic way of segmenting an image on the fly as soon as the image is acquired. This process requires providing necessary accuracy to be useful in the production environment.

In the last years, different methods have been proposed to tackle the problem of creating convolutional neural networks (CNN) designed to be an efficient architecture for pixel-wise semantic segmentation. This networks can produce a segmentation map for an entire input image in a single forward pass. One of the most successful state-of-the-art deep learning method is based on the Fully Convolutional Networks (FCN) \cite{long2015fully}. The main idea of this approach is to use CNN as a powerful feature extractor by replacing the fully connected layers by convolution one to output spatial feature maps instead of classification scores. Those maps are further upsampled to produce dense pixel-wise output. Moreover, this approach achieved an improvement in segmentation accuracy over common methods on standard datasets like PASCAL VOC \cite{everingham2015pascal}. This method has been further improved and now known as U-Net neural network \cite{ronneberger2015u}. The U-Net architecture uses skip connections to combine low-level feature maps with higher-level ones, which enables precise pixel-level localization. A large number of feature channels in upsampling part allows propagating context information to higher resolution layers. This type of network architecture proven themselves in binary image segmentation competitions for satellite image analyses \cite{iglovikov2017satellite, iglovikov2018ternausnet, iglovikov2018ternausnetv2, zhang2018road, zhong2016fully} and other applications \cite{shvets2018automatic, shvets2018angiodysplasia}.

\section{Dataset}
\begin{figure}[t]
\begin{center}
\includegraphics[width=8cm]{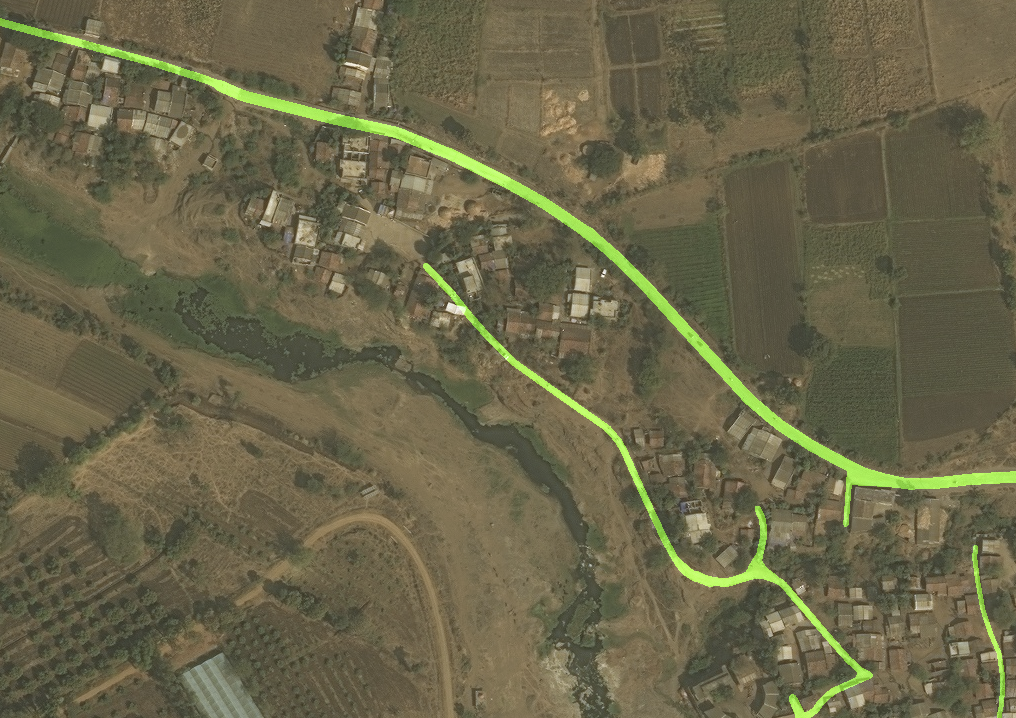}
\end{center}
   \caption{A satellite image with overlay binary masks where green pixels indicate class membership (roads).}
\label{fig:road}
\end{figure}

The training data for road extraction challenge contains 6226 satellite imagery in RGB format. Each image has the size of 1024x1024 pixels. These images have 50cm pixel resolution, collected by DigitalGlobe's satellite \cite{deepglobe_website, demir2018deepglobe}. Moreover, each image in the training dataset contains a paired mask for road labels (see Fig. \ref{fig:road}). The mask is given in a grayscale format, with white standing for road pixel, and black standing for the background. It is worth to mention that the values of the mask image may not be pure 0 and 255. As a result, the recommended threshold for binarization is 128. The labels are not perfect due to the cost of annotating segmentation mask, especially in rural areas. In addition, small roads within farmlands are not annotated consciously. To measure the performance of our model we were also provided by 1243 validation images that do not contain masks. The predicted masks for the validation images should be upload to deepglobe website \cite{deepglobe_website}. 

\section{Model}
\begin{figure*}
\begin{center}
\includegraphics[width=\textwidth,height=9cm]{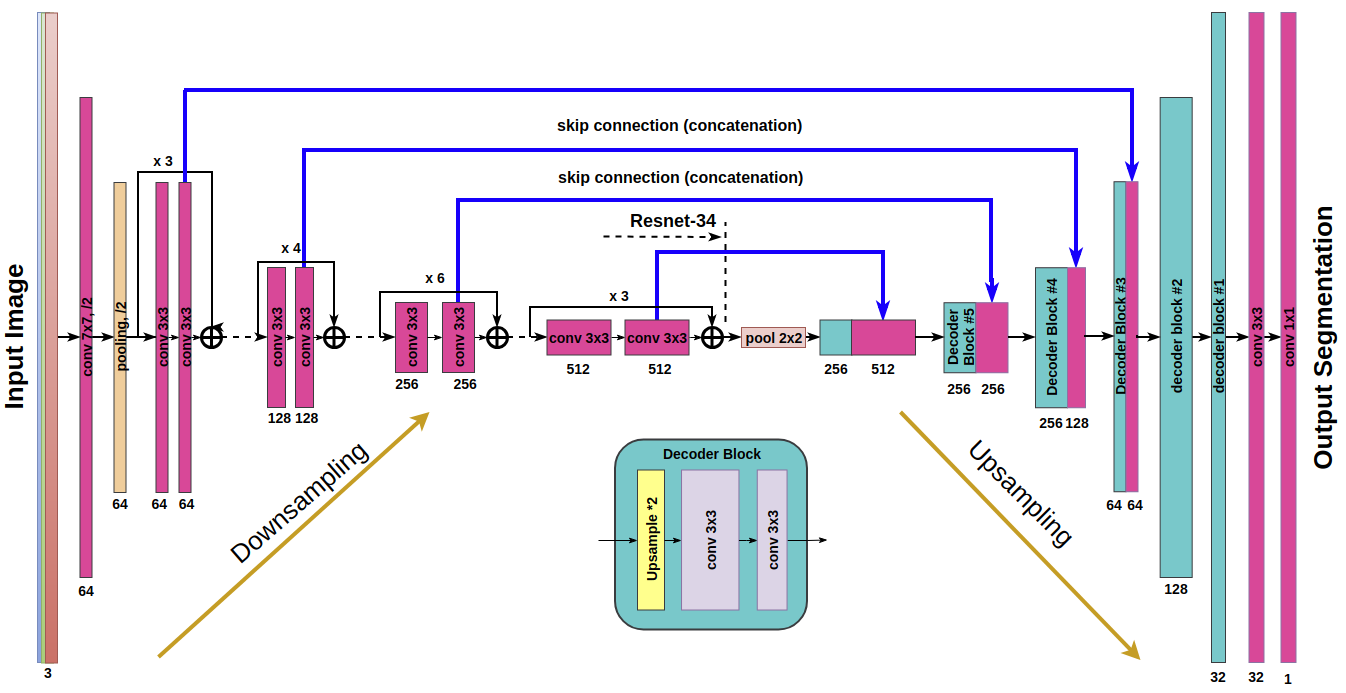}
\end{center}
   \caption{Segmentation networks based on the encoder-decoder architecture of U-Net family. This network uses the pre-trained ResNet-34 network as an encoder. Each box corresponds to a multi-channel feature map. The number of channels is pointed below the box. The height of the box represents a feature map resolution while its thickness is proportional to the number of channels. The blue arrows denote skip-connections where information is transmitted from the encoder to the decoder..}
\label{fig::fpn}
\end{figure*}

Objects segmentation in different scales is challenging in particular for small objects. For this problem, we use a fully convolutional network from U-Net family to implement road segmentation. In general, a U-Net-like architecture consists of a contracting path to capture context and of a symmetrically expanding path that enables precise localization (for example, see Fig.\ref{fig::fpn}). The contracting path follows the typical architecture of a convolutional network with alternating convolution and pooling operations and progressively downsamples feature maps, increasing the number of feature maps per layer at the same time. Every step in the expansive path consists of an upsampling of the feature map followed by a convolution. Hence, the expansive branch increases the resolution of the output. To localize, upsampled features, the expansive path combines them with high-resolution features from the contracting path via skip-connections \cite{ronneberger2015u}. The output of the model is a pixel-by-pixel mask that shows the class of each pixel. We use a slightly modified version of the original U-Net model that previously proved itself very useful for segmentation problems with limited amounts of data, for example, see \cite{iglovikov2018ternausnet}.

As an improvement over the standard U-Net architecture, we use similar networks with pre-trained encoders. Our network has a U-Net-like architecture that uses pre-trained ResNet-34 \cite{he2016deep} networks as an encoder (see Fig. \ref{fig::fpn}). The encoder starts with the initial block that performs convolution with a kernel of size $7\times7$ and stride $2$. This block is followed by max-pooling with stride $2$. The later portion of the network consists of repetitive residual blocks. In every residual block, the first convolution operation is implemented with stride $2$ to provide downsampling, while the rest convolution operations use stride $1$. In addition, the decoder of the network consists of several decoder blocks that are connected with the corresponding encoder block. As for vanilla U-Net, the transmitted block from the encoder is concatenated to the corresponding decoder block. Each decoder block includes $1\times1$ convolution operation that reduces the number of filters by $4$, followed by batch normalization and transposed convolution to upsample the feature map.

\section{Training}
We use Jaccard index (Intersection Over Union) as the evaluation metric. It can be interpreted as a similarity measure between a finite number of sets. For two sets $A$ and $B$, it can be defined as following:
\begin{equation}
\label{jaccard_iou}
    J(A, B) = \frac{|A\cap B|}{|A\cup B|} = \frac{|A\cap B|}{|A|+|B|-|A\cap B|}
\end{equation}
Since an image consists of pixels, the last expression can be adapted for discrete objects in the following way:
\begin{equation}
\label{dicrjacc}
J=\frac{1}{n}\sum\limits_{i=1}^n\left(\frac{y_i\hat{y}_i}{y_{i}+\hat{y}_i-y_i\hat{y}_i}\right)
\end{equation}
where $y_i$ and $\hat{y}_i$ are a binary value (label) and a predicted probability for the pixel $i$, correspondingly.

Since image segmentation task can also be considered as a pixel classification problem, we additionally use common classification loss functions, denoted as $H$. For a binary segmentation problem, $H$ is a binary cross entropy, while for a multi-class segmentation problem $H$ is a categorical cross entropy.

The final expression for the generalized loss function is obtained by combining (\ref{dicrjacc}) and $H$ as following: 
\begin{equation}
\label{free_en}
L=\alpha H+(1-\alpha) (1-J)
\end{equation}
Minimizing this loss function, we simultaneously maximize probabilities for right pixels to be predicted and maximize the intersection $J$ between masks and corresponding predictions. The weighted parameter $\alpha=0.7$ is found from the evaluation of the network using hold out data set.  

For training our network, we split our dataset using 1/4 hold out values for validation. Then, on the fly, we make several augmentations to increase the train size artificially. For spatial augmentation, we use scale transformations $0.6-1.4$ of the original image and mask. Then, we randomly rotate the image and mask by 30 degrees. From the resulting image and mask, we take random crops with size 448x448 pixels. These images are subject to color transformation such as random contrast/brightness/HSV. One video card GTX1080$Ti$ with 11 Gb of memory allows using the batch size of 8 images.  

We train our network using Adam optimizer with learning rate 1e-4 and decay 1e-4 \cite{kingma2014adam}. The training is done for 20k iterations (batches) saving weights from several best iterations. Since the data set is fairly limited in its size and labeling of train images is not robust the predicted value for IoU is varied significantly between iterations. To reduce the effect of over-fitting, we used spatial dropout operation on the output of our network with $p=0.3$. 

We made predictions on the whole image with 1024x1024 pixels without padding because the side is divisible by $32=2^5$. To improve the robustness of our predictions, we also implemented test time augmentation (TTA) that composed of averaging of 4 predictions that correspond to 90 degrees rotation each. 

\section{Conclusions}
We developed a binary segmentation model using the encoder-decoder network with skip connections. For this network, we used encoder based on the pre-trained ResNet-34 network while the decoder was similar to a vanilla U-Net decoder. In addition, we neatly designed loss function that simultaneously takes into account binary cross entropy and intersection over union (IoU). To improve the performance of our method we also used test time augmentation technique. The best public score of our model on the public leaderboard is 0.64. This method can be further improved implementing cross-validation for five folds, improving image augmentation or making more TTA transformations. Last but not least, to make robust and precise predictions of our method the easiest way is to prepare labeled masks in high quality. After that our method could potentially be optimized to work on embedded devices and to provide real-time road extraction. 

\section*{Acknowledgment}
The authors would like to thank Open Data Science community \cite{ods_website} for many valuable discussions and educational help in the growing field of machine/deep learning.

{\small
\bibliographystyle{ieee}
\bibliography{landlib}

\begin{thebibliography}{10}\itemsep=-1pt

\bibitem{deepglobe_website}
\url{http://deepglobe.org/}.

\bibitem{ods_website}
\url{http://ods.ai/}.

\bibitem{demir2018deepglobe}
I.~Demir, K.~Koperski, D.~Lindenbaum, G.~Pang, J.~Huang, S.~Basu, F.~Hughes,
  D.~Tuia, and R.~Raskar.
\newblock Deepglobe 2018: A challenge to parse the earth through satellite
  images.
\newblock {\em arXiv preprint arXiv:1805.06561}, 2018.

\bibitem{everingham2015pascal}
M.~Everingham, S.~A. Eslami, L.~Van~Gool, C.~K. Williams, J.~Winn, and
  A.~Zisserman.
\newblock The pascal visual object classes challenge: A retrospective.
\newblock {\em International journal of computer vision}, 111(1):98--136, 2015.

\bibitem{he2016deep}
K.~He, X.~Zhang, S.~Ren, and J.~Sun.
\newblock Deep residual learning for image recognition.
\newblock In {\em Proceedings of the IEEE conference on computer vision and
  pattern recognition}, pages 770--778, 2016.

\bibitem{iglovikov2017satellite}
V.~Iglovikov, S.~Mushinskiy, and V.~Osin.
\newblock Satellite imagery feature detection using deep convolutional neural
  network: A kaggle competition.
\newblock {\em arXiv preprint arXiv:1706.06169}, 2017.

\bibitem{iglovikov2018ternausnet}
V.~Iglovikov and A.~Shvets.
\newblock Ternausnet: U-net with vgg11 encoder pre-trained on imagenet for
  image segmentation.
\newblock {\em arXiv preprint arXiv:1801.05746}, 2018.

\bibitem{iglovikov2018ternausnetv2}
V.~I. Iglovikov, S.~Seferbekov, A.~V. Buslaev, and A.~Shvets.
\newblock Ternausnetv2: Fully convolutional network for instance segmentation.
\newblock {\em arXiv preprint arXiv:1806.00844}, 2018.

\bibitem{kingma2014adam}
D.~P. Kingma and J.~Ba.
\newblock Adam: A method for stochastic optimization.
\newblock {\em arXiv preprint arXiv:1412.6980}, 2014.

\bibitem{long2015fully}
J.~Long, E.~Shelhamer, and T.~Darrell.
\newblock Fully convolutional networks for semantic segmentation.
\newblock In {\em Proceedings of the IEEE conference on computer vision and
  pattern recognition}, pages 3431--3440, 2015.

\bibitem{ronneberger2015u}
O.~Ronneberger, P.~Fischer, and T.~Brox.
\newblock U-net: Convolutional networks for biomedical image segmentation.
\newblock In {\em International Conference on Medical image computing and
  computer-assisted intervention}, pages 234--241, 2015.

\bibitem{shvets2018angiodysplasia}
A.~Shvets, V.~Iglovikov, A.~Rakhlin, and A.~A. Kalinin.
\newblock Angiodysplasia detection and localization using deep convolutional
  neural networks.
\newblock {\em arXiv preprint arXiv:1804.08024}, 2018.

\bibitem{shvets2018automatic}
A.~Shvets, A.~Rakhlin, A.~A. Kalinin, and V.~Iglovikov.
\newblock Automatic instrument segmentation in robot-assisted surgery using
  deep learning.
\newblock {\em arXiv preprint arXiv:1803.01207}, 2018.

\bibitem{wang2016review}
W.~Wang, N.~Yang, Y.~Zhang, F.~Wang, T.~Cao, and P.~Eklund.
\newblock A review of road extraction from remote sensing images.
\newblock {\em Journal of Traffic and Transportation Engineering (English
  Edition)}, 3(3):271--282, 2016.

\bibitem{zhang2018road}
Z.~Zhang, Q.~Liu, and Y.~Wang.
\newblock Road extraction by deep residual u-net.
\newblock {\em IEEE Geoscience and Remote Sensing Letters}, 2018.

\bibitem{zhong2016fully}
Z.~Zhong, J.~Li, W.~Cui, and H.~Jiang.
\newblock Fully convolutional networks for building and road extraction:
  preliminary results.
\newblock In {\em Geoscience and Remote Sensing Symposium (IGARSS), 2016 IEEE
  International}, pages 1591--1594. IEEE, 2016.

\end{thebibliography}
}

\end{document}